\pdfoutput=1

\documentclass[11pt]{article}

\usepackage{EMNLP2023}

\usepackage{times}
\usepackage{latexsym}

\usepackage[T1]{fontenc}

\usepackage[utf8]{inputenc}

\usepackage{microtype}

\usepackage{inconsolata}
\usepackage{bbm}
\usepackage{smile}
\usepackage{graphicx}
\usepackage{algorithm2e}
\usepackage{booktabs}

\title{Better Explain Transformers by Illuminating Important Information}

\author{
   Linxin Song$^{1,4}$, \ Yan Cui$^2$, \ Ao Luo$^1$, \ Freddy Lecue$^3$ \and Irene Li$^{4,5}$ \\
   $^1$Waseda University \quad $^2$Kyoto University \quad $^3$INRIA \quad $^4$University of Tokyo  \quad $^5$Smartor.me, Inc\\
   {\normalsize  \{songlx.imse.gt@ruri, luo.ao@toki\}.waseda.jp,} \\
   {\normalsize yancui@kuicr.kyoto-u.ac.jp, \ freddy.lecue@inria.fr, \ ireneli@ds.itc.u-tokyo.ac.jp}
}

\begin{document}

\maketitle

\begin{abstract}
Transformer-based models excel in various natural language processing (NLP) tasks, attracting countless efforts to explain their inner workings. Prior methods explain Transformers by focusing on the raw gradient and attention as token attribution scores, where non-relevant information is often considered during explanation computation, resulting in confusing results.
In this work, we propose highlighting the important information and eliminating irrelevant information by a refined information flow on top of the layer-wise relevance propagation (LRP) method. Specifically, we consider identifying syntactic and positional heads as important attention heads and focus on the relevance obtained from these important heads.
Experimental results demonstrate that irrelevant information does distort output attribution scores and then should be masked during explanation computation. Compared to eight baselines on both classification and question-answering datasets, our method consistently outperforms with over 3\% to 33\% improvement on explanation metrics, providing superior explanation performance. 
Our anonymous code repository is available at: \url{https://github.com/LinxinS97/Mask-LRP} 

\end{abstract}

\section{Introduction}


Transformer~\cite{vaswani2017attention} currently serves as the fundamental structure for state-of-the-art models~\cite{kenton2019bert, radford2019language, liu2020roberta, llama, llama2}.
The power of these models provides convincing results in multiple Natural Language Processing (NLP) tasks.
However, building a robust Transformer-based model to assist trustworthy human decision-making processes requires an understanding of the internal mechanisms of the Transformers~\cite{kovaleva2019revealing, jain2019attention, qiang2022counterfactual}.

In NLP tasks, tokens are prevalently utilized to signify a word or a fragment of a word (also known as a \textit{subword}), serving as the input for Transformers.
To comprehend the influence of input tokens on a Transformer, helping us to understand which part of input the Transformer is most interested in, a typical approach involves determining the \textit{attribution score} of input tokens by leveraging the information captured by the attention matrix obtained from each attention head~\cite{bach2015pixel, attgrad, partiallrp, chefer2021transformer, gae}. 
A high attribution score signifies that the input token likely plays a pivotal role in the model's decision-making process for a specific class, output word, or answer index.

To derive attribution scores for each input token, recent approaches utilized information within a trained Transformer, such as input-gradients~\cite{pmlr-v70-shrikumar17a, Ancona2019gradient}, raw attention matrices~\cite{rollout} or the combination of input-gradients and attention matrix~\cite{attgrad, qiang2022attcat}.
The underlying premise for those methods is that input token gradients reflect the token's significance during backpropagation, while attention mechanisms capture the between-token interactions.
However, both theoretical and empirical results~\cite{chefer2021transformer, qiang2022attcat, ali2022xai} indicate that not all types of information embedded within the gradient and attention mechanisms contribute towards the explanations. 
They either fail to or can only partially aid in understanding which token primarily contributes to the Transformer's decision-making process.

To solve this issue, we follow the line of work known as Layer-wise Relevance Propagation (LRP, \citet{bach2015pixel}) with refined information flow to derive compelling attribution scores for each token.
The information flow within LRP parameterized by each attention head mirrors that of the Transformer, concentrating on distinct portions of the input tokens, and attention heads focusing on irrelevant information can disrupt this flow, causing explanation confusion.
We refine the information flow within LRP by illuminating the attention head that focuses on important information and reducing the attention head that zeroes in on less important information.

To achieve this, we illuminate the important attention head by adopting a head mask generated from dataset statistics.
We first label the attention heads concentrating on a specific syntactic relationship as \textit{syntactic} attention heads. 
Syntactic relations (e.g., nominal subject) are extensively utilized to define the relations between tokens in NLP~\cite{partiallrp}, which establish a directional relation between two words.
Furthermore, we designate the attention head that predominantly centers on a fixed relative position as a \textit{positional} attention head, which reflects the internal feature (e.g., spatial position) of token embedding.
We encapsulate \textit{syntactic} and \textit{positional} within a head mask, which we use to refine the information flow during the LRP process.
To further reduce the irrelevant information, we obtain the attribution score by rolling out the relevance of the attention head from each attention blocks with the corresponding gradient~\cite{chefer2021transformer}.

To evaluate the performance of our method, we compared it with eight strong baselines across five classification datasets and two question-answering datasets.
The results reveal that our method outperforms others in explanation performance, demonstrating a distinguished capacity to assign influential tokens from both interaction and internal perspectives.
Furthermore, an ablation study uncovers that irrelevant information can obfuscate the LRP process, subsequently leading to a biased explanation of input tokens. 
The key contributions of our work can be summarized as follows: 
\begin{enumerate}
    \item We refine the information flow within the LRP process by illuminating two types of important information.
    \item Through experiments, we demonstrated that irrelevant information hampers the LRP process. 
    \item Compared to previous state-of-the-art methods, our approach significantly improves explanation performance, achieving over 3.56\% improvement in AOPC and LOdds for classification tasks and 33.02\% for Precision@20 in question answering tasks.
\end{enumerate}

\section{Related Works}
To explain a Transformer in NLP tasks, one common approach involves providing a post-hoc interpretable description of the Transformer's behavior. This approach assists users in understanding which input tokens most significantly influence the model's decision-making process. \citet{rollout} achieve this by leveraging the attention heads for defining more elaborate explanation mechanisms, while \citet{wallace-etal-2019-allennlp} and \citet{atanasova-etal-2020-diagnostic} accomplish this by involving the Integrated Gradients or Input Gradients. Numerous models and domains have employed gradient methods such as Saliency Maps~\cite{zhou2016learning, attgrad}, Gradient$\times$Input~\cite{pmlr-v70-shrikumar17a, NEURIPS2019_80537a94, NEURIPS2021_a284df11, qiang2022attcat}, or Guided Backpropagation~\cite{zeiler2014visualizing}, and these methods have also been effectively transposed and applied to Transformers. 

Concurrently, there have been several attempts to implement Layer-Wise Relevance Propagation (LRP, \citet{bach2015pixel}) in Transformers~\cite{partiallrp,ali2022xai} and other attention-based models~\cite{ding-etal-2017-visualizing}. LRP has been used to explain predictions of diverse models on NLP tasks, including BERT~\cite{kenton2019bert}. Other methodologies for LRP / gradient propagation in Transformer blocks can be found in~\cite{chefer2021transformer, gae}, where the relevance scores are determined by combining attention scores with LRP or attention gradients.

Additionally, a few instances exist where perturbation-based methods have employed input reductions~\cite{feng-etal-2018-pathologies, prabhakaran-etal-2019-perturbation}, aiming to identify the most relevant parts of the input by observing changes in model confidence or leveraging Shapley values~\cite{NIPS2017_8a20a862, atanasova-etal-2020-diagnostic}. Furthermore, a line of work using tensor decomposition to decompose the attention matrix for a faithful Transformer explanation~\cite{kobayashi-etal-2020-attention, kobayashi-etal-2021-incorporating, modarressi-etal-2022-globenc, ferrando-etal-2022-measuring}.

\section{Preliminary}

\subsection{Problem Formulation}

This work focuses on post-hoc explanations of Transformer-based models, like BERT~\cite{kenton2019bert,liu2020roberta} and GPT~\cite{radford2019language}, across various NLP tasks. 
Given a dataset $D$ with each input $\bx_i$ consisting of $T$ tokens, we use a fine-tuned Transformer-based language model, $f(\cdot; \btheta)$, composed of $B$ self-attention blocks with $M$ attention heads each.
We extract each model layer's output for analysis, with layer input denoted as $\bx^{(n)}$ and $n$ ranging from $1$ to $N$. 
Here, $\bx^{(N)}$ and $\bx^{(1)}$ signify the model input and output, respectively, as information propagation starts from the output to the input.

We aim to understand the attribution of input $\bx^{(N)} \in D$ to the output $\bx^{(1)} \in \{c_1...c_K\}$ ($K$ denoting classification task classes or question answering task tokens). 
We seek an attribution function $\bR^{(N)} = R(\bx^{(N)})$ evaluating each token's contribution to output $\bx^{(N)}$. 
An ideal $\bR^{(N)}$ assigns high attribution scores to influential tokens, causing output confidence to flatten or predictions to flip when these tokens are removed or masked.

\subsection{Layer-wise Relevance Propagation}

The Layer-wise Relevance Propagation (LRP,~\citet{bach2015pixel}) is used to compute the attribution score $\bR^{(N)}$ of each input token, propagating relevance from the predicted class or index backward to the input tokens.

The LRP applies the chain rule to propagate gradients with respect to the output $\bx^{(1)}$ at index $c$, denoted as $\bx^{(1)}_c$:
\begin{equation}
\nabla \bx_j^{(n)} = \frac{\partial\bx^{(1)}_c}{\partial\bx_j^{(n)}} = \sum_i \frac{\partial\bx^{(1)}_c}{\partial \bx^{(n-1)}_i} \frac{\partial \bx^{(n-1)}_i}{\partial \bx^{(n)}_j},
\end{equation}
where $j$ and $i$ are element indices in $\bx^{(n)}$ and $\bx^{(n-1)}$ respectively.
The layer operation on two tensors $\bX$ and $\bY$ is denoted as $L^{(n)}$, typically indicating the input feature map and weights for layer $n$. 
The relevance propagation follows the Deep Taylor Decomposition~\cite{montavon2017explaining}:
\begin{align}
\label{eq:taylor_decomp}
R_j^{(n)} &= \cG(\bX, \bY, \bR^{(n-1)}) \\
&= \sum_i \bX_j\frac{\partial L_i^{(n)}(\bX, \bY)}{\partial \bX_j} \frac{R_i^{(n-1)}}{L_i^{(n)}(\bX, \bY)} , \notag
\end{align}
with $j$ and $i$ denoting elements in $R^{(n)}$ and $R^{(n-1)}$ respectively. 
This equation obeys the conservation rule:
\begin{equation}
\sum_j R_j^{(n)} = \sum_i R_i^{(n-1)}.
\end{equation}
We begin relevance propagation with $R^{(0)}$ as a one-hot vector indicating the target class or index $c\in\bx^{(1)}$.

LRP presumes non-negative activation functions and is incompatible with functions outputting both positive and negative values, like GELU~\cite{hendrycks2016gaussian}. 
As \citet{chefer2021transformer} done, we overcome this by filtering out negative values and selecting the positive subset of indices $q=\{(i,j)|x_i w_{ij} \geq 0\}$ for relevance propagation:
\begin{align}
R_j^{(n)} &= \cG(x, w, q, R^{(n-1)}) \notag \\
&= \sum_{\{i|(i,j)\in q\}} \frac{x_j w_{ji}}{\sum\limits_{\{j'|(j', i)\in q\}} x_{j'} w_{j'i}} R_i^{(n-1)}.
\end{align}

\section{Layer-wise Relevance Propagation Through Important Attention Head}

In this work, we empirically show that irrelevant information can detrimentally impact the LRP process. 
Therefore, our focus should be directed toward the important information while concurrently eliminating irrelevant information within the LRP process.
In this section, we initially classify two kinds of important information (Sec.\ref{sec:info}), followed by introducing the method to extract this information in each layer (Sec.\ref{sec:mask}). 
Subsequently, we illustrate the technique of concentrating on the important information extracted during Layer-wise Relevance Propagation (LRP, Sec.~\ref{sec:mlrp}).

\subsection{Important Information Flows in Transformer}
\label{sec:info}
Understanding Transformer-based models in NLP tasks entails grasping the important information each attention head prioritizes. This information in an input sentence comprises internal and interaction information~\cite{partiallrp, qiang2022attcat}. 
Interaction information explores if Transformer's encoder heads focus on tokens tied to core syntactic relationships, while internal information refers to an input where an attention head focuses on a fixed position for token embedding~\cite{partiallrp}. 
In this work, to capture the above types of information, we identify two functions that attention heads might be playing: (1) syntactic: the head points to tokens in a specific syntactic relation, and (2) positional: the head points to a specific relative position.
Not all syntactic relations are suitable for defining the core component of a sentence. \citet{de2014universal} classifies the syntactic relations into nominal, clauses, modifier words, and function words. While nominal (subject, object) and modifier words (adverb, adjectival modifier) are frequent, others like vocatives (common in conversations), expletives (e.g., "it" and "their" in English), and dislocated elements (frequent in Japanese) don't define a sentence's core and explain on them can confuse human understanding.
Therefore, we identify four core syntactic relations: nominal subject (\textit{nsubj}), direct object (\textit{dobj}), adjectival modifier (\textit{amod}), and adverbial modifier (\textit{advmod}), which contains the core information of a whole sentence. 
The selected syntactic relations establish directional links between two words or linguistic units.
For example, in "\textit{The car is red}", \textit{car} is the nsubj target for \textit{red}. 
Hence, in LRP, important information the relevance contains of a layer input $\bx^{(n_b)}$ in the self-attention block $b$ at layer $n_b$ can be decomposed as:
\begin{equation}
\label{eq:imp_info}
\bR_{\text{imp}}^{(n_b)} = \bR_{\text{synt}}^{(n_b)} + \bR_{\text{pos}}^{(n_b)},
\end{equation}
where $\bR_{\text{imp}}^{(n_b)}$ denotes the important information, $\bR_{\text{synt}}^{(n_b)}$ and $\bR_{\text{pos}}^{(n_b)}$ the information from syntactic relations and relative positions, respectively. 
The next section will detail preserving important information in the LRP process by identifying the important attention heads.

\subsection{Identifying Important Heads}
\label{sec:mask}
\begin{figure}
    \centering
    \includegraphics[width=\linewidth]{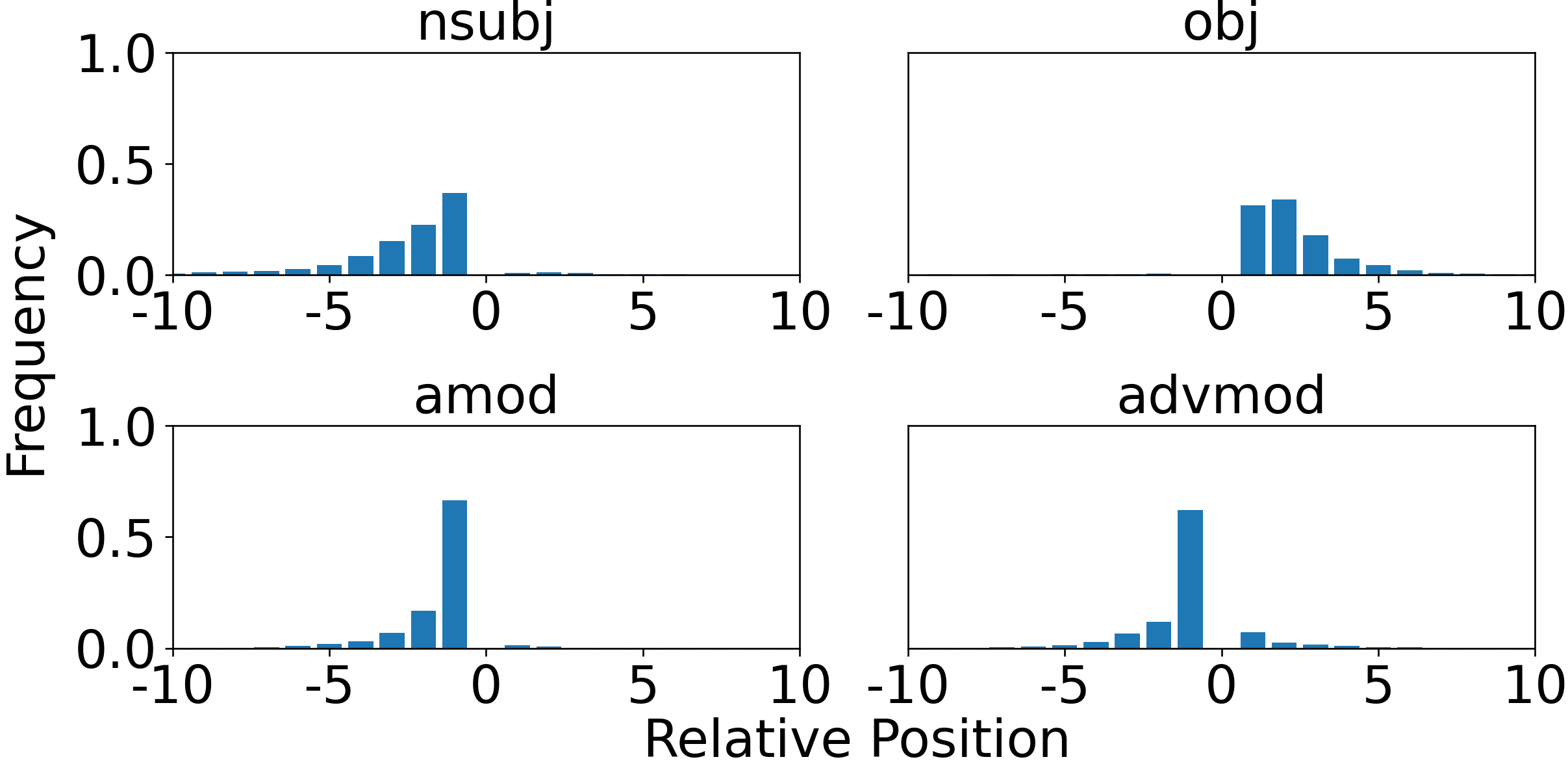}
    \caption{Distributions of the relative positions dependent for different syntactic relations in SST2.}
    \label{fig:sst2_deprel_stats}
\end{figure}
\begin{figure*}
    \centering
    \includegraphics[width=\linewidth]{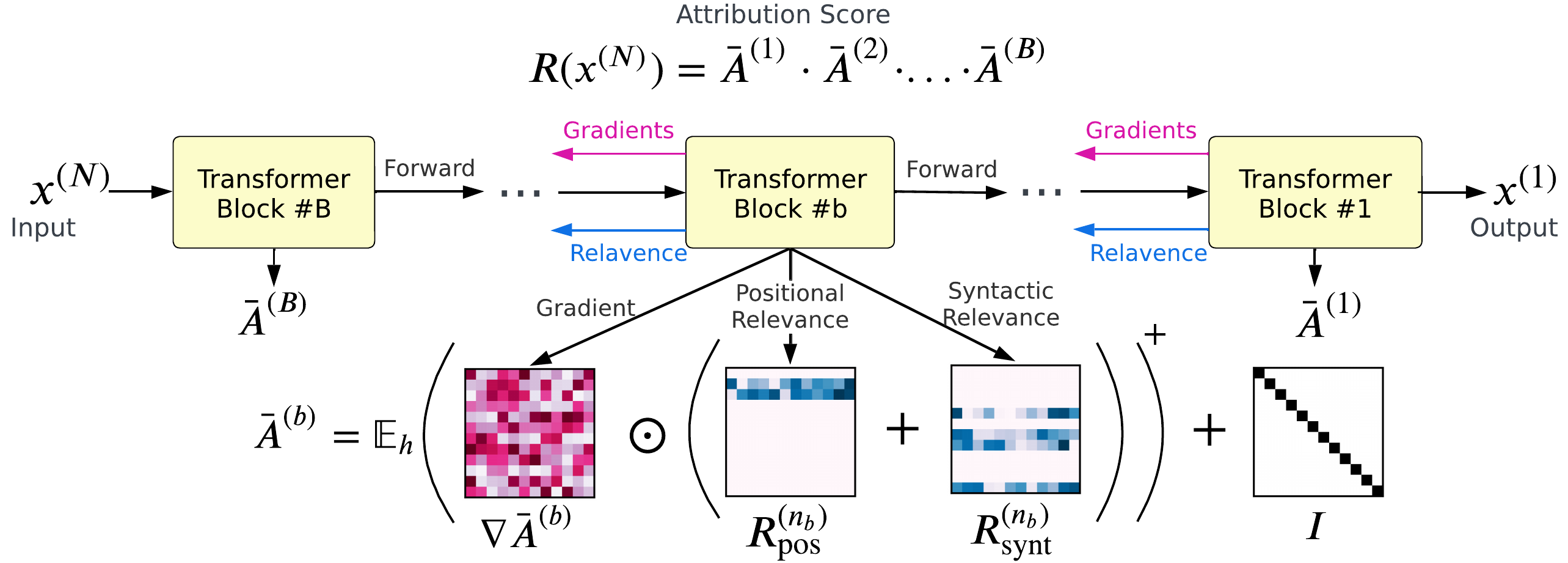}
    \caption{Illustration of our method. Gradients and relevance are propagated through the Transformer block from the final layer to the first layer. We extract two types of important information during the LRP process in all blocks by identifying the important heads.}
    \label{fig:important_head_lrp}
\end{figure*}
To illuminate the influence of the attention heads that are oriented towards important information, we create a head mask denoted as $\cM\in\RR^{B\times M}$ by combining two separate masks: $\cM_{\text{synt}}$ and $\cM_{\text{pos}}$.
The mask $\cM$ is constructed as follows:
\begin{equation}
\label{eq:mask}
    \cM = \cM_{\text{synt}} + \cM_{\text{pos}}.
\end{equation}
$\cM_{\text{synt}}$ represents the syntactic mask generated based on the statistical analysis of syntactic relations within each text, while the positional mask $\cM_{\text{pos}}$ is derived from the positional analysis of the specific Transformer-based model chosen for the study.

\paragraph{Syntactic mask.} 
We first obtain the distribution of the $k$-th syntactic relation at each token position, denoted as $\blambda_k$. 
Here, $\lambda_{k}^i$ represents the probability of the $k$-th syntactic relation appearing at position $i$ (as depicted in Fig.~\ref{fig:sst2_deprel_stats}).
The attention head mask for syntactic relations, denoted as $\cM^{(b,m)}_{\text{synt}}$, can be derived as follows:
\begin{equation}
\label{eq:synt_mask}
\cM^{(b,m)}_{\text{synt}} = \sum_{k\in K}\mathbbm{1}_{\left\{\alpha_k^{(b,m)} > \text{max}(\blambda_k) + \xi_{\text{synt}}\right\}},
\end{equation}
where $K = \{\text{nsubj, dobj, amod, advmod}\}$ represents the set of core syntactic relations, $\alpha_k^{(b,m)}\in[0, 1]$ denotes the frequency of the $m$-th attention head at block $b$ assigning its highest attention weight to the $k$-th syntactic relation. 
The threshold $\xi_{\text{synt}}$ determines the level of probability at which an attention head is considered syntactic relation-specific.
In this work, we set $\xi_{\text{synt}} = 0.1$ to ensure that the selected attention head is not solely focused on a specific token position but exhibits a substantial probability of capturing syntactic relations.

\paragraph{Positional mask.} 
We also examine attention heads that exhibit a high degree of focus on specific relative positions (e.g., $...,-1, +1, +2,...$).
We refer to these attention heads as "positional" if, most of the time, their maximum attention weight is assigned to a specific relative position.
To identify these attention heads, we utilize a positional mask denoted as $\cM^{(b,m)}_{\text{pos}}$, which collects the indices of attention heads that satisfy the positional criteria. The positional mask is defined as follows:
\begin{equation}
\label{eq:pos_mask}
\cM^{(b,m)}_{\text{pos}} = \sum_{i\in I}\mathbbm{1}_{\left\{\alpha_i^{(b,m)} > \xi_{\text{pos}}\right\}},
\end{equation}
where $\alpha_i^{(b,m)}\in[0, 1]$ denotes the frequency of the $m$-th attention head at block $b$ assigning its highest attention weight to the $i$-th relative position, $I=\{...,-1,+1,...\}$ denotes the set of relative positions and $\xi_{\text{pos}}$ is set to 0.8, as previously mentioned, to ensure that we capture attention heads primarily focusing on the positional information.

\subsection{Layer-wise Relevance Propagation Through Important Heads}
\label{sec:mlrp}
To gain deeper insights into the important information within the Transformer model, we specifically focus on the Layer-wise Relevance Propagation (LRP) process between important attention heads across different layers and obtain the final attribution score. 
The process of our proposed method is illustrated in Fig.~\ref{fig:important_head_lrp}.

According to the type of information a relevance contains, the relevance of each attention head in the self-attention block at layer $n_b$ can be defined as a combination of two types of relevance w.r.t. attention heads: important relevance and irrelevant relevance.
Recalling the Eq.~\eqref{eq:taylor_decomp} and \eqref{eq:imp_info}, we have:

{\small\begin{align}
\bR^{(n_b)} &= \cG\left(\bX, \bY, \bR_{\text{imp}}^{(n_b-1)} + \bR_{\text{others}}^{(n_b-1)}\right) \notag \\
&= \cG\left(\bX, \bY, \bR_{\text{synt}}^{(n_b-1)} + \bR_{\text{pos}}^{(n_b-1)} + \bR_{\text{others}}^{(n_b-1)}\right),
\end{align}}
in each Transformer block.
Here, $\bR_{\text{others}}^{(n_b-1)}$ corresponds to the relevance output from attention heads that are not specific to important information.
To highlight the important relevance $\bR_{\text{imp}}^{(n_b-1)}$ in the LRP process, we employ the $b$-th block's mask $\cM^{(b)}$ obtaining from Eq.~\eqref{eq:mask}:

{\small
\begin{align*}
\bR^{(n_b)} := \bR_{\text{synt}}^{(n_b)} + \bR_{\text{pos}}^{(n_b)} = \cG(\bX, \bY, \cM^{(b)}\bR^{(n_b-1)}).
\end{align*}}

To keep the conservation after adopting the mask, we apply normalization to $\bR_{\text{synt}}^{(n_b)}$ and $\bR_{\text{pos}}^{(n_b)}$ as follows:
\begin{align*}
\bR_{\text{synt}}^{(n_b)} &:= \bR_{\text{synt}}^{(n_b)}\frac{\left|\sum R_{\text{synt}}^{(n_b)}\right|}{\left|\sum R^{(n_b)}\right|}\cdot \frac{\sum R^{(n_b-1)}}{\sum R_{\text{synt}}^{(n_b)}}, \\
\bR_{\text{pos}}^{(n_b)} &:= \bR_{\text{pos}}^{(n_b)}\frac{\left|\sum R_{\text{pos}}^{(n_b)}\right|}{\left|\sum R^{(n_b)}\right|}\cdot \frac{\sum R^{(n_b-1)}}{\sum R_{\text{pos}}^{(n_b)}}.
\end{align*}
The normalization step ensures the conservation rule is maintained, i.e., $\sum R_{\text{synt}}^{(n_b)} + \sum R_{\text{pos}}^{(n_b)} = \sum R^{(n_b-1)}$. Note that we have omitted the subscript of the index (e.g., $i$, $j$) to enhance readability.

We output the final attribution $\bR^{(N)}$ by leveraging the rollout of weighted attention relevance~\cite{chefer2021transformer} of each block $b$:

{\small
\begin{align}
    \Bar{\bA}^{(b)} &= \EE_h\left(\nabla\bA^{(b)}\odot \left(\bR_{\text{synt}}^{(n_b)} + \bR_{\text{pos}}^{(n_b)}\right)\right)^+ + I\\
    \bR(x^{(N)})&=\bar{\bA}^{(1)}\cdot \bar{\bA}^{(2)}\cdot ... \cdot \bar{\bA}^{(B)},
\end{align}
}
where $\odot$ denotes the Hadamard product, $\bA^{(b)} = \text{softmax}(\bQ^{(b)}\cdot\bK^{(b)\top} / \sqrt{d_h})$ is the attention matrix obtain from query $\bQ$ and key $\bK$ in block $b$, and $\nabla \bA^{(b)}$ denotes the corresponding gradient. We use the superscript $a^+$ to denote the operation $max(0, a)$.

\section{Experiment}
\subsection{Experiment Setup}

\paragraph{Implementation details.}
For the classification task, we use pretrained $\text{BERT}_{\text{base}}$~\cite{kenton2019bert} with a 512 token input limit and attribute the [CLS] token as the classifier input.
For question answering, we compare our method with three baselines using pretrained $\text{BERT}_{\text{base}}$, GPT-2~\cite{radford2019language}, and RoBERTa~\cite{liu2020roberta}, assessing the effect of model scale and tokenizer on information flow. We evaluate the attribution of the start and end answer indices.

Our model-agnostic method can apply to various Transformer-based models with minimal modifications. 
We obtain all results from the validation set across all methods, focusing on the post-hoc explanation with fixed model parameters. 
Variance is limited to the baseline using a randomly generated mask.

\begin{table*}[t!]
\centering
\resizebox{\linewidth}{!}{%
\begin{tabular}{lcccccccccc}
\toprule
\multirow{2}{*}{\textbf{Methods}} &
  \multicolumn{2}{c}{\textbf{SST-2}} &
  \multicolumn{2}{c}{\textbf{IMDB}} &
  \multicolumn{2}{c}{\textbf{Yelp}} &
  \multicolumn{2}{c}{\textbf{MNLI}} &
  \multicolumn{2}{c}{\textbf{QQP}} \\ 
  \cmidrule(r){2-3} 
  \cmidrule(r){4-5}
  \cmidrule(r){6-7}
  \cmidrule(r){8-9} 
  \cmidrule(r){10-11} 
 &
  AOPC $\uparrow$ &
  LOdds  $\downarrow$ &
  AOPC  $\uparrow$ &
  LOdds    $\downarrow$ &
  AOPC     $\uparrow$ &
  LOdds   $\downarrow$ &
  AOPC   $\uparrow$ &
  LOdds  $\downarrow$ &
  AOPC  $\uparrow$ &
  LOdds $\downarrow$ \\ \midrule
RawAtt &
  0.374 &
  -0.992 &
  0.354 &
  -1.593 &
  0.376 &
  -1.513 &
  0.135 &
  -0.399 &
  0.447 &
  -5.828 \\
Rollout &
  0.337 &
  -0.911 &
  0.334 &
  -1.456 &
  0.244 &
  -0.770 &
  0.137 &
  -0.396 &
  0.437 &
  -5.489 \\ \midrule
LRP &
  0.336 &
  -0.888 &
  0.288 &
  -1.271 &
  0.163 &
  -0.464 &
  0.131 &
  -0.395 &
  0.438 &
  -5.745 \\
PartialLRP &
  0.396 &
  -1.052 &
  0.370 &
  -1.726 &
  0.401 &
  -1.688 &
  0.136 &
  -0.401 &
  0.445 &
  -5.718 \\
GAE &
  0.423 &
  -1.171 &
  0.384 &
  -1.853 &
  0.404 &
  -1.682 &
  0.144 &
  -0.421 &
  0.447 &
  -5.923 \\ \midrule
CAM &
  0.399 &
  -1.086 &
  0.365 &
  -1.883 &
  0.298 &
  -1.473 &
  0.132 &
  -0.386 &
  0.450 &
  -5.988 \\
GradCAM &
  0.341 &
  -0.855 &
  0.236 &
  -0.974 &
  0.104 &
  -0.229 &
  0.126 &
  -0.369 &
  0.449 &
  -5.953 \\
AttCAT &
  0.405 &
  -1.110 &
  0.340 &
  -1.697 &
  0.397 &
  \textbf{-2.034} &
  0.138 &
  -0.419 &
  0.447 &
  -5.897 \\ \midrule
Random &
  0.432{\scriptsize$\pm$.005} &
  -1.205{\scriptsize$\pm$.004} &
  0.387{\scriptsize$\pm$.004} &
  -1.898{\scriptsize$\pm$.003} &
  0.426{\scriptsize$\pm$.005} &
  -1.886{\scriptsize$\pm$.007} &
  0.142{\scriptsize$\pm$.002} &
  -0.415{\scriptsize$\pm$.021} &
  0.448{\scriptsize$\pm$.001} &
  -5.998{\scriptsize$\pm$.012} \\
Ours &
  \textbf{0.438} &
  \textbf{-1.208} &
  \textbf{0.392} &
  \textbf{-1.906} &
  \textbf{0.434} &
  -1.898 &
  \textbf{0.148} &
  \textbf{-0.445} &
  \textbf{0.451} &
  \textbf{-6.001} \\ \bottomrule
\end{tabular}%
}
\caption{AOPC and LOdds results of all methods in explaining $\text{BERT}_{\text{base}}$ model on each dataset. The best results are marked in bold. Note that a method with high AOPC and low LOdds is desirable, indicating a strong ability to mark influential tokens. The results of the Random mask are average and standard deviation between five runs. We also provide the comparison with SOTA tensor decomposition method in Appendix \ref{apdx:td_comp}.}
\label{tab:main_res}
\end{table*}

\paragraph{Datasets.}
We choose the validation set on seven datasets across the sentiment classification: SST-2~\cite{sst2}, IMDB~\cite{imdb}, Yelp Polarity~\cite{yelp}, duplicated question classification: QQP~\cite{qqp}, natural language inference: MNLI~\cite{mnli} and question answering: SQuADv1~\cite{squadv1} and SQuADv2~\cite{squadv2} to evaluate all methods.
SST-2, IMDB, and Yelp Polarity take a single sentence as input, while QQP and MNLI use a pair of sentences for their target.
Specifically, we extract the data marked as \textit{duplicate} (with ground truth label $1$) in QQP for evaluation. Details of the model and datasets are in Appendix~\ref{apdx:ext_imp}.

\paragraph{Evaluation metrics.}
We use AOPC and LOdds for classification evaluation, and precision@20 for question-answering evaluation.
To evaluate post-hoc explanation interpretability in a classification task, we measure model confidence for a specific class before and after masking influential tokens, using both linear (AOPC) and non-linear (LOdds) metrics~\cite{qiang2022attcat}. 
AOPC and LOdds aim to detect the change of confidence before and after the influential tokens are removed, which are formularized as:
\begin{equation}
\text{AOPC}(k) = \frac{1}{T}\sum_{t=1}^T f_{\hat{y}}\left(\bx_i;\btheta\right) - f_{\hat{y}}\left(\Tilde{\bx}i^k;\btheta\right),
\end{equation}
\begin{equation}
\text{LOdds}(k) = \frac{1}{T}\sum_{t=1}^T \log{\frac{f\left(\Tilde{\bx}_i^k;\btheta\right)}{f\left(\bx_i;\btheta\right)}},
\end{equation}
where $\Tilde{\bx}_i^{k}$ denotes the top-$k\%$ masked input tokens ranked by the attribution score $R(\bx_i^{(N)})$. $f_{\hat{y}}(\cdot; \btheta)$ denotes the model's max confidence w.r.t label $\hat{y}$. 
Furthermore, we use precision@20 to evaluate the question answering task (SQuADv1 and SQuADv2). In QA tasks, precision@20 will not introduce bias because it will not remove the ground truth answer from the input, and the model that has a low precision@20 means that the model cannot capture a correct mapping between the answer part and the ground truth index.

\paragraph{Hyperparameters}
In this work, we use two hyperparameters: $\xi_{\text{synt}}$ and $\xi_{\text{pos}}$ for the corresponding masks.
As we mentioned in the main context, we choose $0.1$ for $\xi_{\text{synt}}$ and $0.8$ for $\xi_{\text{pos}}$.
One reason why we choose these values is that we empirically found that the highest frequency for the syntactic relations is almost lower than $0.7$ for a specific relative position.
Therefore, $\xi_{\text{synt}} = 0.1$ ensure the syntactic mask effectively filters out the attention head, which is focusing on irrelevant information, or just focusing on a specific position, and $\xi_{\text{pos}}=0.8$ help us to capture the rest attention heads that are focusing mainly on a specific relative position, which is filtered by the syntactic mask.
Although the two masks are complementary, many attention heads still focus on various relative positions so that we cannot identify their function and mark them as irrelevant attention heads.
 
\subsection{Baselines}
\label{sec:baselines}

We categorize eight baselines into three groups based on their characteristics with one additional random baseline:
\paragraph{Attention maps}: \textbf{RawAtt}~\cite{rollout} uses the mean attention weights from the final Transformer block as attribution scores, while \textbf{Rollout}~\cite{rollout} rolls out average attention weights from all Transformer blocks.
\paragraph{Relevance-based}: \textbf{LRP}~\cite{bach2015pixel} uses output-to-input layer relevance as attribution scores. \textbf{PartialLRP}~\cite{partiallrp} calculates relevance at the model's final layer. \textbf{GAE}~\cite{gae} propagates attention gradients to the final layer to obtain attribution scores.
\paragraph{Gradient-based}: \textbf{CAM}~\cite{zhou2016learning} and \textbf{GradCAM}~\cite{attgrad} use the final layer gradient and its weighted version by final layer attention respectively as attribution scores. \textbf{AttCAT}~\cite{qiang2022attcat} combines the summation of attention weight from each Transformer block with input gradient.

In addition, we include \textbf{Random}, a baseline using a randomly generated mask (maintaining the same mask rate, i.e., $\|\cM_{\text{random}}\|=\|\cM_{\text{ours}}\|$, as our method) to show that our method effectively identifies the crucial head in the Transformer model.
\begin{table}[t!]
\centering
\resizebox{\linewidth}{!}{%
\begin{tabular}{ccccccc}
\toprule
\multirow{2}{*}{\textbf{Method}} & \multicolumn{3}{c}{\textbf{SQuADv1}}  & \multicolumn{3}{c}{\textbf{SQuADv2}}   \\ \cmidrule(r){2-4}\cmidrule(r){5-7}
                        & $\text{BERT}_{\text{base}}$  & GPT-2 & RoBERTa & $\text{BERT}_{\text{base}}$ & GPT-2 & RoBERTa \\ \midrule
Rollout                 & 4.62                         & 5.86      & 8.04   & 6.15     & 5.54                   & 5.87        \\
RawAtt                  & 36.33                        & 28.97     & 45.61  & 4.69     &  27.85                & 18.09        \\
AttCAT                  & 31.44                        & 17.53     & 47.32   & 18.81     & 16.99                & 23.39        \\ \midrule
Ours                    & \textbf{52.97}               &  \textbf{51.62}  & \textbf{67.31} & \textbf{27.03}     & \textbf{49.63}  & \textbf{56.41}        \\ \bottomrule
\end{tabular}%
}
\caption{Precision@20 results of the selected explanation methods on SQuAD datasets. Higher Precision@20 is better, indicating the marked influential tokens highly overlap with the answer text.}
\label{tab:qa}
\end{table}

\subsection{Results}
\label{sec:main}

\begin{figure}[t!]
    \centering
    \includegraphics[width=\linewidth]{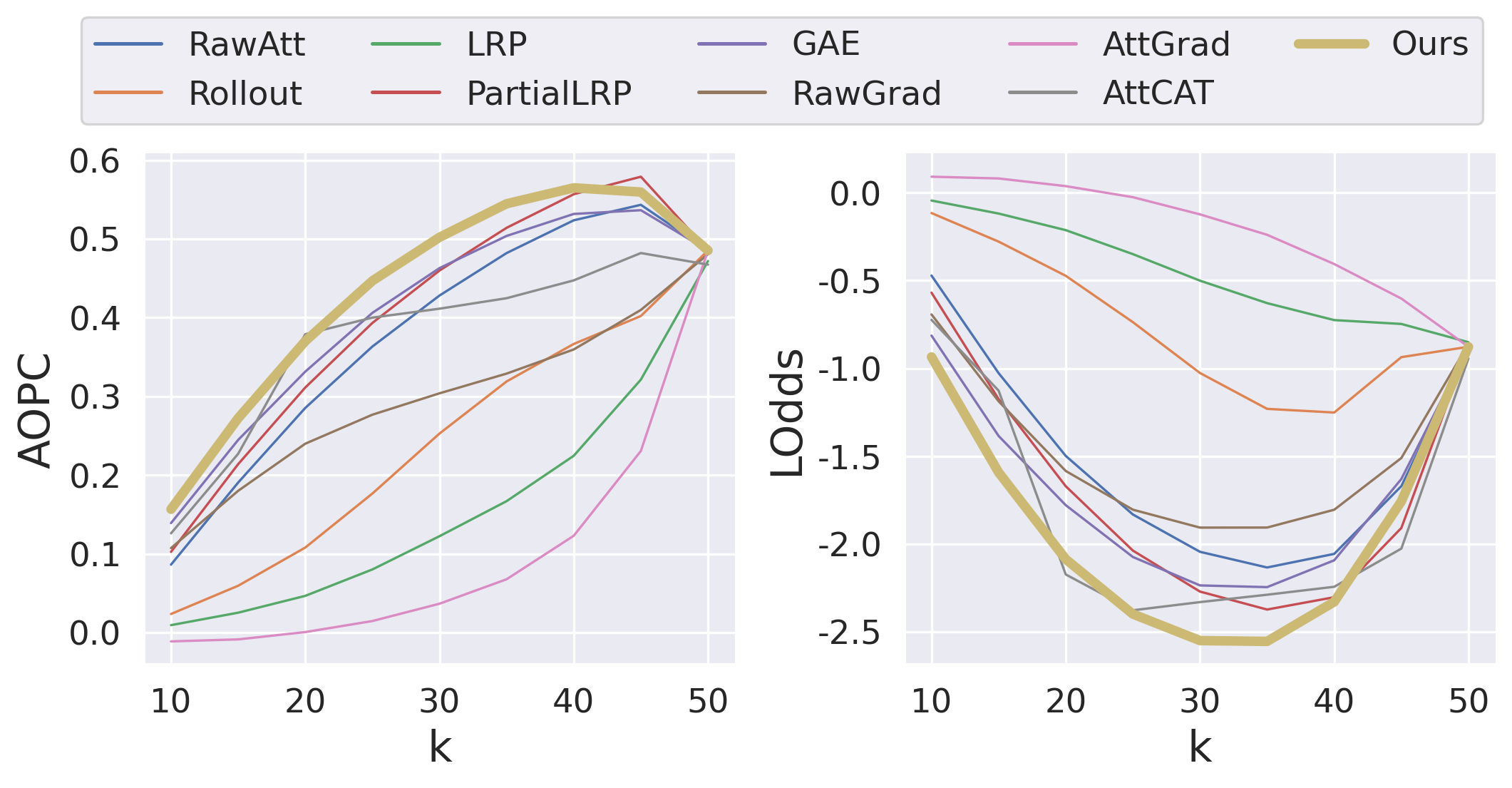}
    \caption{AOPC and LOdds scores of different methods in explaining $\text{BERT}_{\text{base}}$ against the corruption rate $k$ on SST-2. Note that higher AOPC and lower LOdds scores are better.}
    \label{fig:remove_rate_res}
\end{figure}

We assessed the explanation performance of each method within classification tasks by computing mean AOPC and LOdds across five benchmark datasets, detailed in Tab.\ref{tab:main_res}. 
Remarkably, the performance across all post-hoc explanation methods remained stable, independent of random initialization, except for a randomly initialized mask method. 
Our approach generally surpassed others, achieving the highest AOPC and lowest LOdds, indicating superior accuracy in identifying influential tokens. Fig.\ref{fig:remove_rate_res} displays performance curves against pruning rate $k$, endorsing our method's performance at every rate. It consistently outperformed gradient-based methods, particularly in handling lengthy token lists. Attention information from larger matrices often includes irrelevant details that assign high attribution to non-influential tokens, reducing the quality of explanations (see Sec.\ref{sec:abla} for more). For the question-answering task, we evaluated Precision@20 on two SQuAD datasets. As per Tab.\ref{tab:qa}, our method consistently outperformed the baselines, demonstrating accurate attribution to influential answer tokens.

\begin{figure}[t!]
    \centering
    \includegraphics[width=\linewidth]{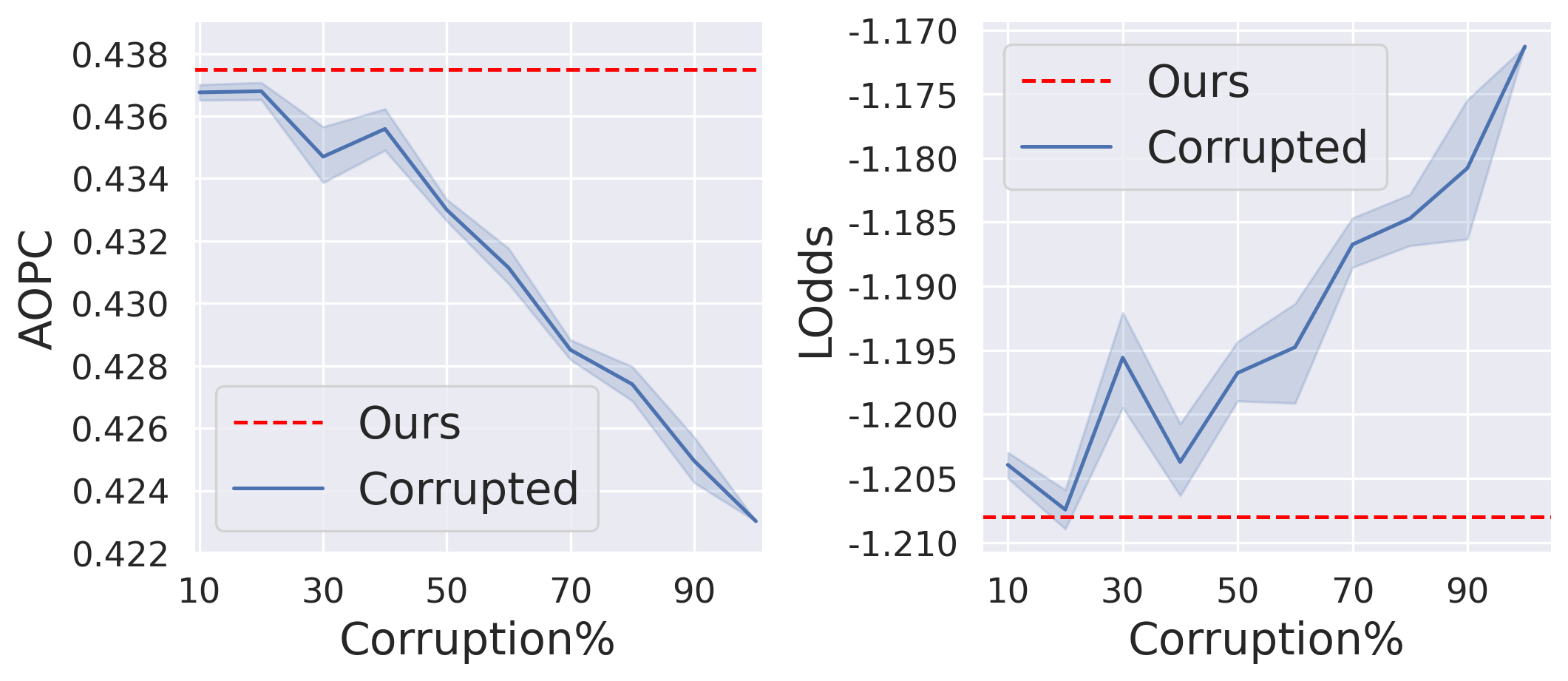}
    \caption{Comparison before and after corrupting the generated mask on SST-2. The blue line combines the solid line (average values) and shadow areas (standard deviation). The method's ability to explain becomes dropped after adding corruption.}
    \label{fig:abla_sst2}
\end{figure}
\begin{figure*}[t!]
    \centering
    \includegraphics[width=\linewidth]{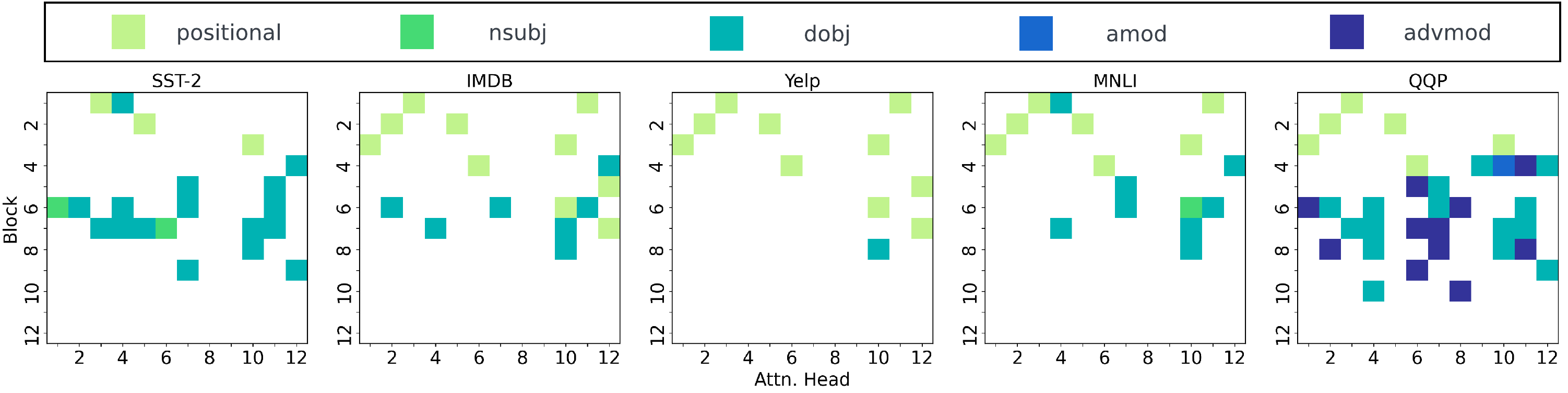}
    \caption{Different types of important heads in $\text{BERT}_{\text{base}}$ model cross different dataset. The $x$-axis denotes the position of the attention head, while the $y$-axis is the position of the Transformer block. It is obvious that attention heads in previous blocks tend to focus on simple internal information (e.g., position), while attention heads in later blocks tend to focus on the complex interactions between tokens (e.g., syntactic relations).}
    \label{fig:mask_visualization}
\end{figure*}
\begin{figure}[t!]
    \centering
    \includegraphics[width=\linewidth]{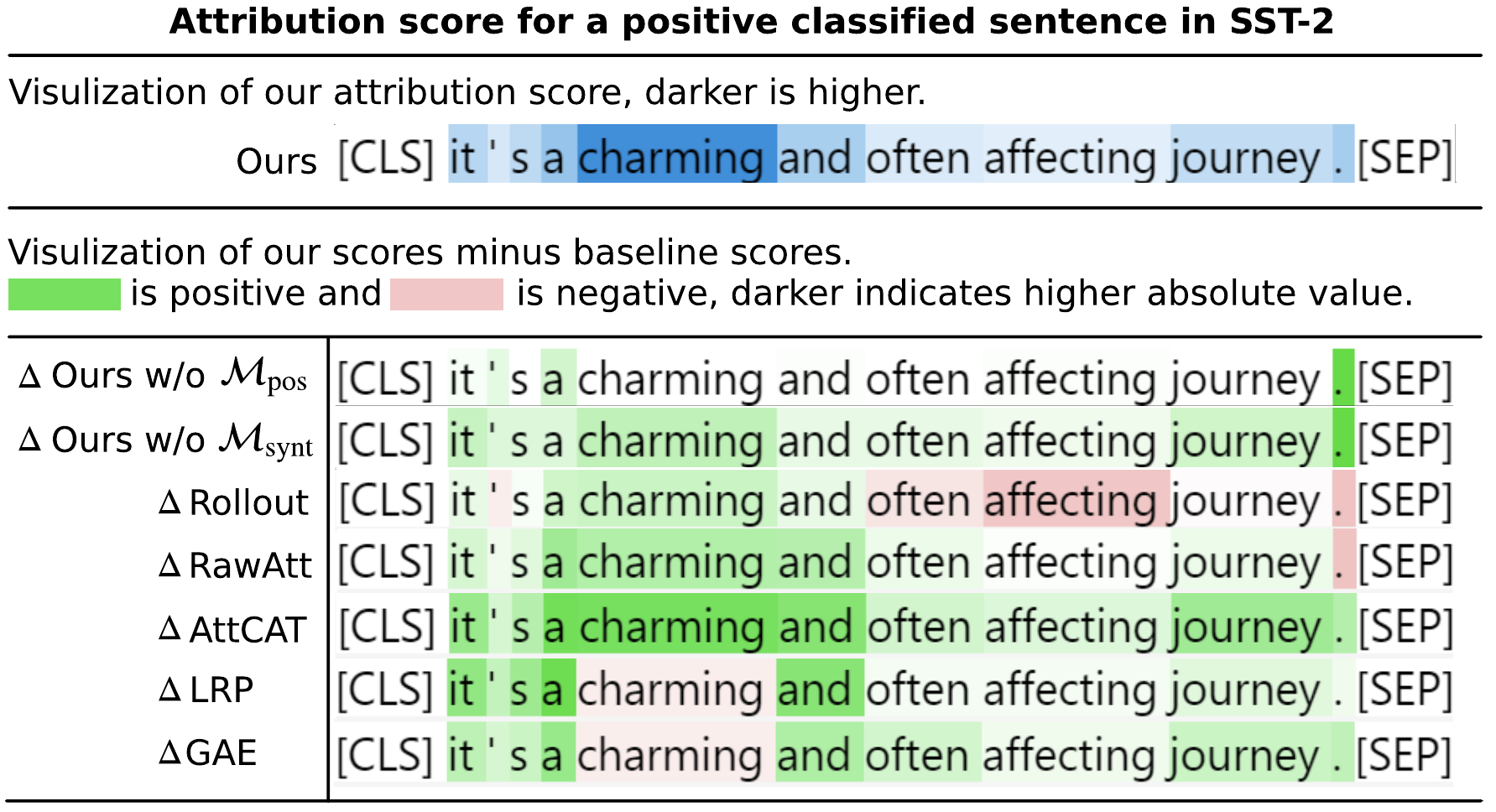}
    \caption{
    The comparison of attribution scores between our method (shown in the first line) and baselines on a positive classified sentence.
    Tokens highlighted in green represent those receiving more attention from our method than the baseline, while those in red signify the opposite. Our method emphasizes more on both internal and interaction information.
    We put results of other datasets in Appendix.\ref{apdx:extra_visul}
    }
    \label{fig:visul}
\end{figure}
\subsection{Assessing the Impact of Important and Irrelevant Information}
\label{sec:abla}
In this section, we seek to address two key questions: (1) does our method effectively identify the attention head that focuses on important information? and (2) does the residual, irrelevant information that other heads concentrate on adversely affect the explanation?

To answer the first question, we carry out an ablation study where we replace our mask with a randomly generated mask, maintaining the same mask rate as discussed in Sec.\ref{sec:baselines}, to examine if this alteration impacts the explanatory capacity. 
The results, as reported on line 12 in Tab.\ref{tab:main_res}, clearly demonstrate that our method consistently outperforms the variant with a randomly generated mask. 
This underscores that our method is capable of identifying a set of attention heads that can robustly explain the information flow within a Transformer.

For the second question, we derive our answer by collating findings from Tab.\ref{tab:main_res} and Fig.\ref{fig:abla_sst2}. 
We discover from Tab.~\ref{tab:main_res} that even with a random mask, our method exhibits superior explanation performance than other relevance-based methods such as GAE because of the less focus on irrelevant information. 
This suggests that irrelevant information flow in the Transformer greatly affects the LRP, thereby confusing the explanation of input tokens.
In addition, we conducted another ablation study where we randomly switched a portion of the remaining zeros in $\cM_{\text{ours}}$. These zeros in the mask correspond to the irrelevant information the Transformer focuses on, and their alteration can be interpreted as a corruption of the generated mask. If our method employs a 100\% corrupted mask (a mask filled with ones), it degenerates to GAE.
We observed the variance in explanation performance at different corruption rates (ranging from 10\% to 100\%) on SST-2, the results of which are displayed in Fig.~\ref{fig:abla_sst2}. 
Notably, it is clear that the rate of performance decline is closely related to the corruption rate and ultimately converges to the performance of GAE. This evidence substantiates the notion that irrelevant information can interfere with the LRP process at each layer, thereby resulting in a perplexing explanation.

\subsection{Visualizing and Analyzing Extracted Attention Heads}
\label{sec:analysis_information}
We visualize both $\cM_{\text{synt}}$ and $\cM_{\text{pos}}$ that our method extracted from $\text{BERT}_{\text{base}}$ according to each dataset.
The resulting visualizations are presented in Fig.~\ref{fig:mask_visualization}. 
We discovered that positional attention heads are predominantly concentrated in the earlier blocks, whereas syntactic attention heads tend to gather in the later blocks. 
This observed phenomenon suggests that Transformers initially learn the simplistic internal information and subsequently propagate this internal information to the subsequent layers. 
This aids the attention heads in these later layers in capturing the interaction information between tokens.
Additionally, we found that during model training on more datasets with long input tokens, such as IMDB and Yelp, there are only a few heads with unipolar function, that is, a head focusing solely on a single pattern, and those heads are filtered by our mask.
Yet, as the experiment results in Sec.~\ref{sec:main} illustrate, the attribution scores assigned solely by these heads are representative enough to provide a persuasive explanation. 
This implies that for binary classification tasks, the important information flow can be remarkably simple, even in the context of complex inputs.
We also examine the explanation performance differences when using $\cM$ compared to solely utilizing $\cM_{\text{pos}}$ or $\cM_{\text{synt}}$ in Tab.~\ref{tab:mask_abla}.
Interestingly, we discover that eliminating one type of mask doesn't substantially impact the explanation performance. This can be attributed to the fact that a single mask type does not alter the ranking of output attribution but rather enriches its detail. Additional insights are provided in the subsequent paragraph.

To delve deeper into the attributions assigned by these important heads, we visualized the difference in attribution scores allocated by our method and other baseline methods in Fig.~\ref{fig:visul}.
The sentence, randomly selected from the SST-2 dataset and depicted in Fig.\ref{fig:visul}, is annotated with a positive sentiment. 
Compared to attention-based methods (Rollout, RawAtt, AttCAT), our approach de-emphasizes less crucial tokens like \textit{affecting}, emphasizing important ones like \textit{charming}. 
Also, unlike relevance-based methods (LRP, GAE) that overlook \textit{journey}, our method pays attention to it due to its link with \textit{charming} via \textit{and}. Thus, our method successfully extracts interaction information, attributing scores based on both single tokens' internal information and their interplay.

\begin{table}[t!]
\centering
\resizebox{0.75\linewidth}{!}{%
\begin{tabular}{lcccc}
\toprule
\multicolumn{1}{c}{\multirow{2}{*}{\textbf{Method}}} & \multicolumn{2}{c}{\textbf{SST-2}} & \multicolumn{2}{c}{\textbf{QQP}} \\ \cmidrule(r){2-3} \cmidrule(r){4-5} 
\multicolumn{1}{c}{} & AOPC  & LOdds  & AOPC  & LOdds  \\ \midrule
Ours                  & 0.438 & -1.208 & 0.451 & -6.001 \\
Ours w/o $\cM_{\text{pos}}$       & 0.438 & -1.208 & 0.450 & -6.001 \\
Ours w/o $\cM_{\text{synt}}$      & 0.437 & -1.205 & 0.449 & -5.998 \\ \bottomrule
\end{tabular}%
}
\caption{Explanation performance comparison of different masks. Only use $\cM_{\text{pos}}$ or $\cM_{\text{synt}}$ still have strong explanation performance.}
\label{tab:mask_abla}
\vspace{-2mm}
\end{table}
\section{Conclusion}
In this study, we propose that irrelevant information in the gradient and attention hampers the explanation process. To address this, we improve the information flow in the LRP process by masking irrelevant attention heads. By illuminating the important information, we show that explanations become more convincing. Our method outperforms nine baseline methods in classification and question answering tasks, consistently delivering better explanation performance.

\section*{Limitations}
Though our method is model-agnostic, limitations in computational resources prevent us from fully exploring its implications for Large Language Models (LLMs) like LLAMA and LLAMA-2~\cite{llama, llama2}, but we provided the implementation in our repository.
We conjecture that LLMs may learn advanced interaction information surpassing the syntactic relationships we defined. This high-level interaction information could potentially allow LLMs to grasp the interplay between sentences or even broader structures like topics, complementing existing research on Transformers' topic learning capability via self-attention mechanisms~\cite{li2023transformers}. 
Additionally, while we've empirically shown that irrelevant information hinders the LRP process, the origins and contents of this irrelevant information remain obscure. We will delve deeper into the nature of such information in future work.

\bibliography{references}
\bibliographystyle{acl_natbib}

\appendix
\clearpage
\section{Why do we choose nsubj, dobj, amod, and advmod?}
\label{apdx:deprel_expl}
Many syntactic relations exist, but not all are suitable for defining the core component of a sentence. \citet{de2014universal} classifies the syntactic relations into nominals, clauses, modifier words, and function words. While nominals (subject, object) and modifier words (adverb, adjectival modifier) are frequent, others like vocatives (common in conversations), expletives (e.g., "it" and "their" in English), and dislocated elements (frequent in Japanese) don't define a sentence's core and explain on them can confuse human understanding.

\section{Extra experiment comparing with tensor decomposition method}
\label{apdx:td_comp}
We provide the comparison results between ours and the SOTA tensor decomposition method ALTC~\cite{ferrando-etal-2022-measuring} in Table \ref{tab:td_comp}.
\begin{table}[h]
\resizebox{\linewidth}{!}{%
\begin{tabular}{lcccccc}
\toprule
\multirow{2}{*}{Methods} & \multicolumn{2}{c}{SST-2}                                          & \multicolumn{2}{c}{IMDB}                                           & \multicolumn{2}{c}{Yelp}                               \\ \cmidrule(r){2-3}\cmidrule(r){4-5}\cmidrule(r){6-7} 
                         & AOPC $\uparrow$                 & LOdds  $\downarrow$              & AOPC  $\uparrow$                & LOdds    $\downarrow$            & AOPC     $\uparrow$             & LOdds   $\downarrow$ \\ \midrule
ALTC                    & 0.369                           & -0.866                           & 0.342                           & -0.748                           & 0.363                           & -1.428               \\
Ours                     & \textbf{0.438} & \textbf{-1.208} & \textbf{0.392} & \textbf{-1.906} & \textbf{0.434} & \textbf{-1.898}              \\ \bottomrule
\end{tabular}
}
\caption{\label{tab:td_comp}AOPC and LOdds results of ALTC and ours in explaining $\text{BERT}_{\text{base}}$ model on SST-2, IMDB, and Yelp. The best results are marked in bold. Note that a method with high AOPC and low LOdds is desirable, indicating a strong ability to mark influential tokens.}
\end{table}

\section{Extra Implementation Details}
\label{apdx:ext_imp}
\paragraph{Environment}
We run all experiments on the device with the following specs:
\begin{itemize}
    \item System: Ubuntu 20.04.4 LTS
    \item CPU: Intel(R) Xeon(R) Platinum 8368 @ 2.40GHz (36 Cores / 72 Threads)
    \item GPU: NVIDIA A100 SXM4 40GB
    \item Memory: 230GB
\end{itemize}
With the above specs, we can complete the evaluation of one dataset within one hour by adopting the multi-process.

\paragraph{Datasets}
The task, amount of training, validation, and testing set numbers are shown in Tab.~\ref{tab:app_data}. 
Note that the dataset of IMDB and Yelp Polarity does not contain a validation set, so we use the test set for our experiment.
Moreover, in QQP, data points are annotated with a binary label as \textit{duplicated} or \textit{not duplicated}.
If we remove the influential tokens in those data marked as \textit{not duplicated}, the model's prediction does not change because the two questions remain different.
Therefore, we select the data marked as \textit{duplicated} for our experiments to see the changing of the model's prediction from \textit{duplicated} to \textit{not duplicated}.
\begin{table}[ht]
\resizebox{\linewidth}{!}{%
\begin{tabular}{l|c|c|c|c}
\toprule\hline
\multicolumn{1}{c|}{Dataset} & Task & Train & Valid & Test \\ \hline
SST-2                                 & Classification           & 6,920          & 872            & 1,821         \\ \hline
IMDB                                  & Classification           & 25,000         & -              & 25,000        \\ \hline
Yelp Polarity                         & Classification           & 560,000        & -              & 38,000        \\ \hline
QQP                                   & Question Paring           & 363,846        & 40,430         & 390,965       \\ \hline
MNLI                                  & Natural Language Inference           & 392,702        & 20,000         & 20,000        \\ \hline
SQuADv1                               & Question Answering            & 87,599         & 10,570         & 9,533         \\ \hline
SQuADv2                               & Question Answering            & 130,319        & 11,873         & 8,862         \\ \hline\bottomrule
\end{tabular}
}
\caption{Statistics for the benchmark dataset we used in this work. Note that IMDB and Yelp Polarity only contains training and test set.}
\label{tab:app_data}
\end{table}

\paragraph{Models}
In this work, we use different pretrained models archived in Hugging Face~\footnote{\url{https://huggingface.co/}} for each task and modify them to adjust for LRP in our implementation.
The models we use for different tasks are shown in Tab.~\ref{tab:app_model}.
Note that there does not exist GPT-2 model pretrained on SQuADv2, so we adopt the model trained on SQuADv1 for SQuADv2 experiments, which also provides convincing performance.
\begin{table}[ht]
\resizebox{\linewidth}{!}{%
\begin{tabular}{l|c|l}
\toprule\hline
\multicolumn{1}{c|}{Dataset} & Model   & \multicolumn{1}{c}{Huggingface Repo}                   \\ \hline
SST-2                         & $\text{BERT}_{\text{base}}$    & textattack/bert-base-uncased-SST-2                      \\ \hline
IMDB                          & $\text{BERT}_{\text{base}}$    & textattack/bert-base-uncased-imdb                       \\ \hline
Yelp                          & $\text{BERT}_{\text{base}}$    & abriceyhc/bert-base-uncased-yelp\_polarity              \\ \hline
QQP                           & $\text{BERT}_{\text{base}}$     & modeltc/bert-base-uncased-qqp                           \\ \hline
MNLI                          & $\text{BERT}_{\text{base}}$    & textattack/bert-base-uncased-MNLI                       \\ \hline
\multirow{3}{*}{SQuADv1}      & $\text{BERT}_{\text{base}}$    & csarron/bert-base-uncased-squad-v1                      \\ \cline{2-3} 
                              & GPT-2    & anas-awadalla/gpt2-span-head-finetuned-squad            \\ \cline{2-3} 
                              & RoBERTa & thatdramebaazguy/roberta-base-squad                     \\ \hline
\multirow{3}{*}{SQuADv2}      & $\text{BERT}_{\text{base}}$    & ericRosello/bert-base-uncased-finetuned-squad-frozen-v2 \\ \cline{2-3} 
                              & GPT-2    & anas-awadalla/gpt2-span-head-finetuned-squad            \\ \cline{2-3} 
                              & RoBERTa & 21iridescent/roberta-base-finetuned-squad2-lwt          \\ \hline\bottomrule
\end{tabular}
}
\caption{Baseline models of different datasets and their Hugging Face repositories.}
\label{tab:app_model}
\end{table}

\begin{figure*}[t!]
    \centering
    \includegraphics[width=\linewidth]{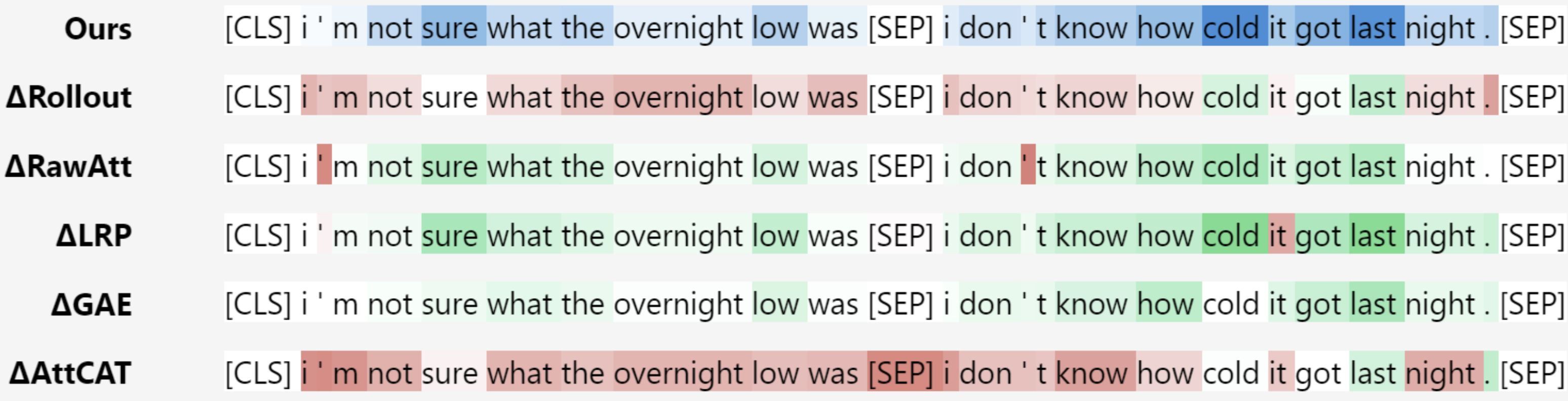}
    \caption{The comparison of attribution scores between our method (shown in the first line) and baselines on an \textbf{entailment} classified sentence pair in MNLI.}
    \label{fig:visu_mnli_0}
\end{figure*}
\begin{figure*}[t!]
    \centering
    \includegraphics[width=\linewidth]{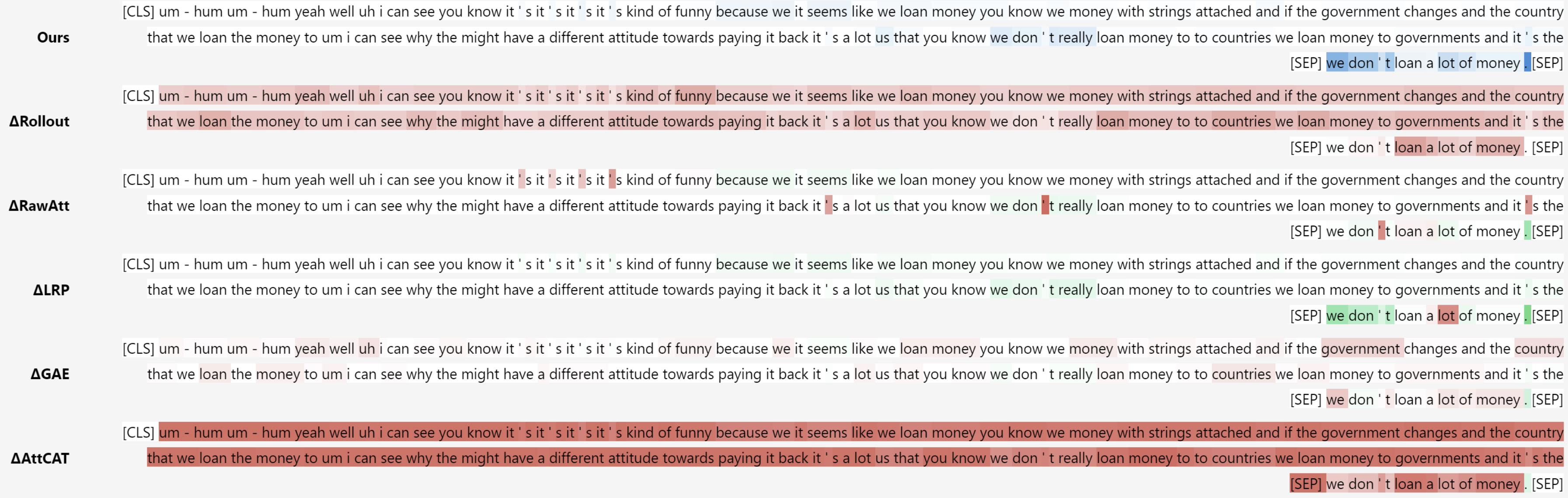}
    \caption{The comparison of attribution scores between our method (shown in the first line) and baselines on a \textbf{neutral} classified sentence pair in MNLI.}
    \label{fig:visu_mnli_1}
\end{figure*}
\begin{figure*}[t!]
    \centering
    \includegraphics[width=\linewidth]{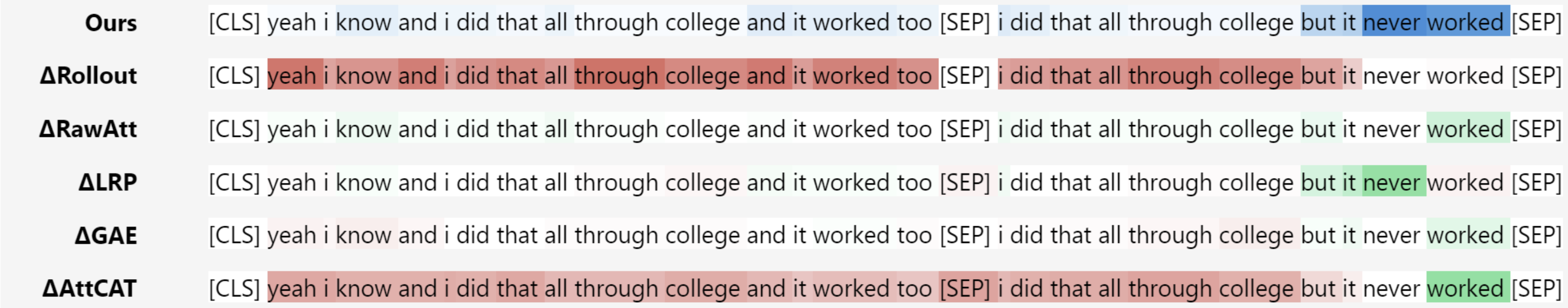}
    \caption{The comparison of attribution scores between our method (shown in the first line) and baselines on a \textbf{contradiction} classified sentence pair in MNLI.}
    \label{fig:visu_mnli_2}
\end{figure*}
\begin{figure*}[t!]
    \centering
    \includegraphics[width=\linewidth]{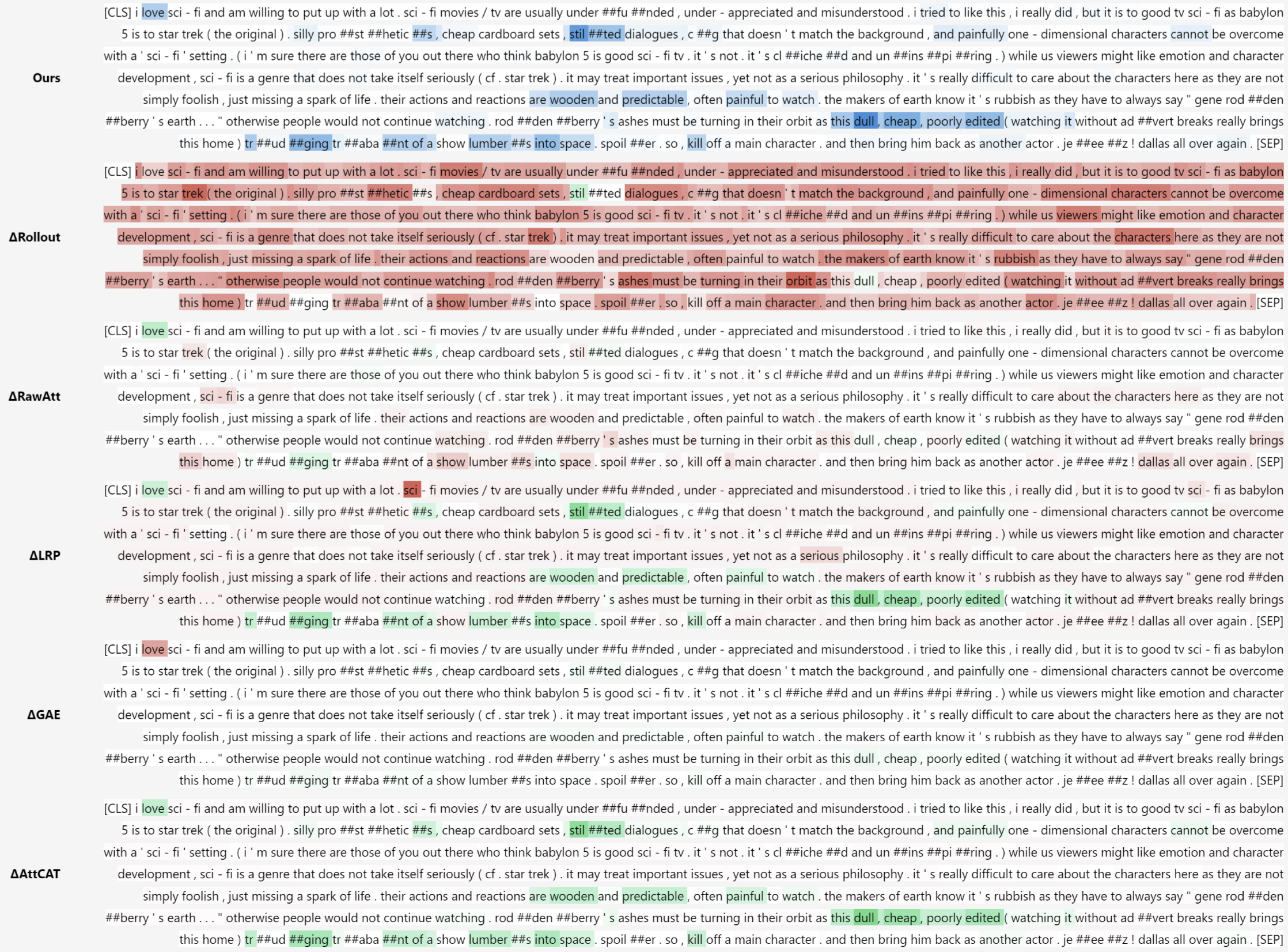}
    \caption{The comparison of attribution scores between our method (shown in the first line) and baselines on a \textbf{negative} classified comment in IMDB.}
    \label{fig:visu_imdb_0}
\end{figure*}
\begin{figure*}[t!]
    \centering
    \includegraphics[width=\linewidth]{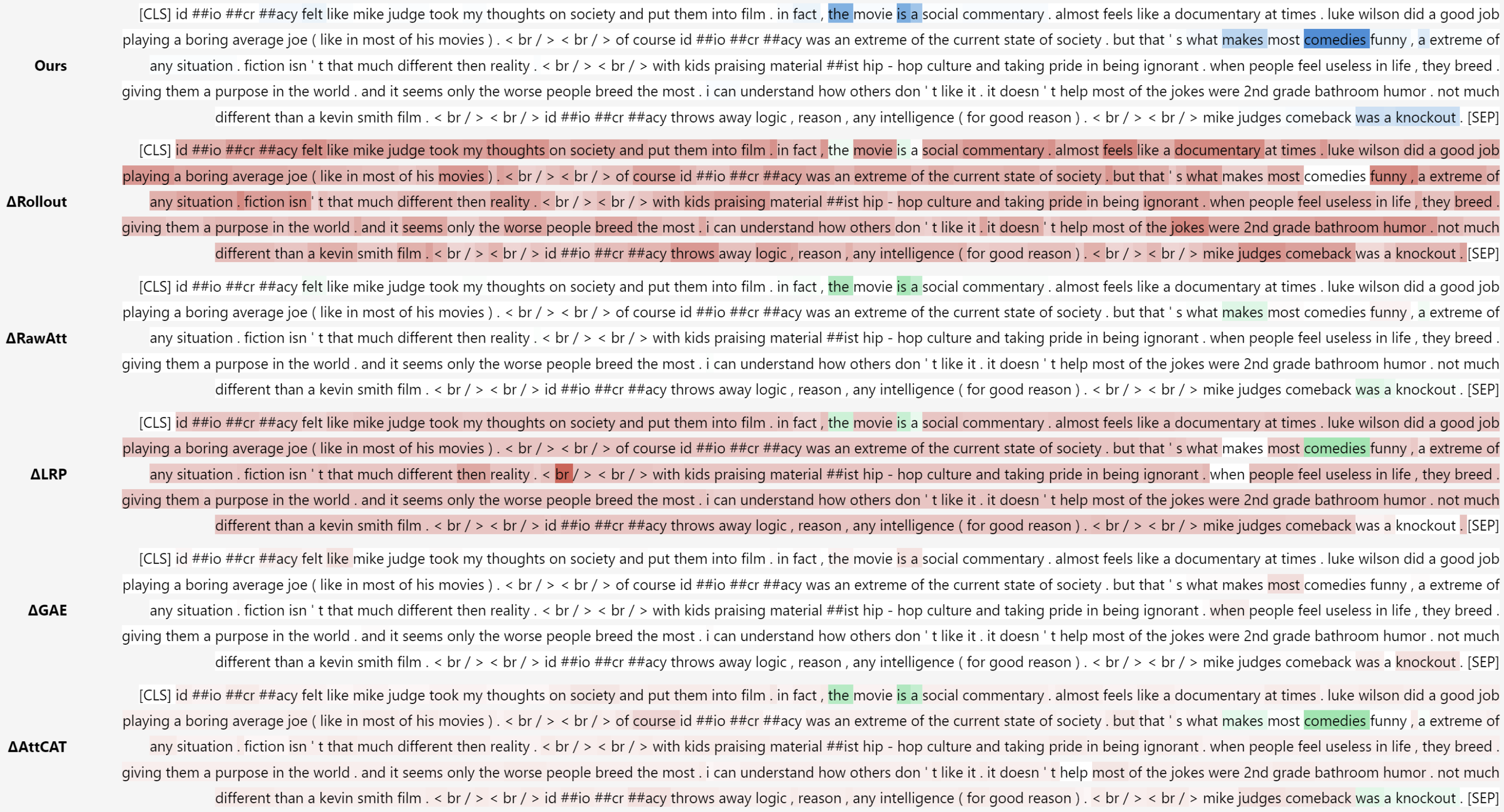}
    \caption{The comparison of attribution scores between our method (shown in the first line) and baselines on a \textbf{positive} classified comment in IMDB.}
    \label{fig:visu_imdb_1}
\end{figure*}
\begin{figure*}[t!]
    \centering
    \includegraphics[width=\linewidth]{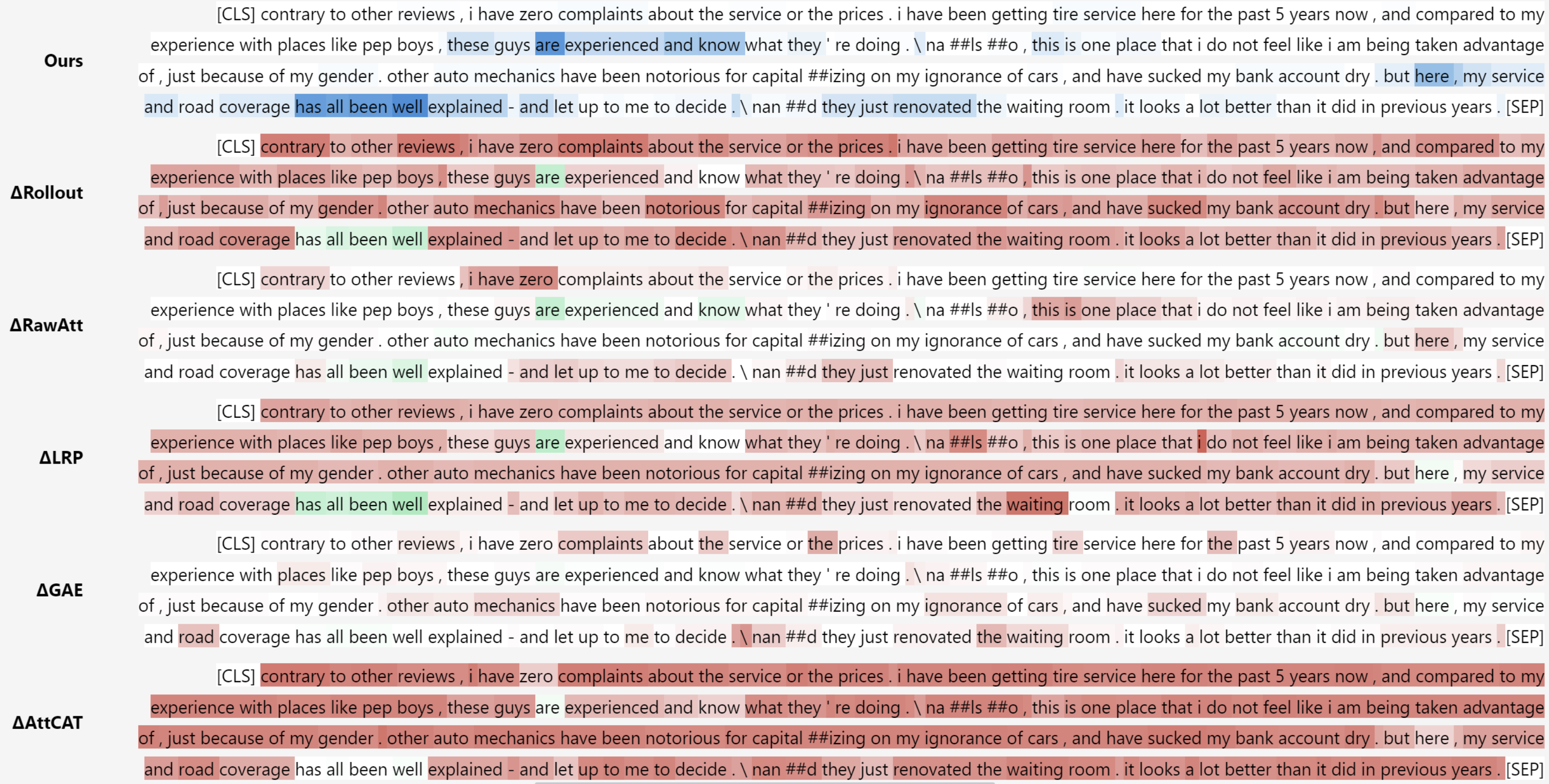}
    \caption{The comparison of attribution scores between our method (shown in the first line) and baselines on a \textbf{negative} classified comment in Yelp Polarity.}
    \label{fig:visu_yelp_0}
\end{figure*}
\begin{figure*}[t!]
    \centering
    \includegraphics[width=\linewidth]{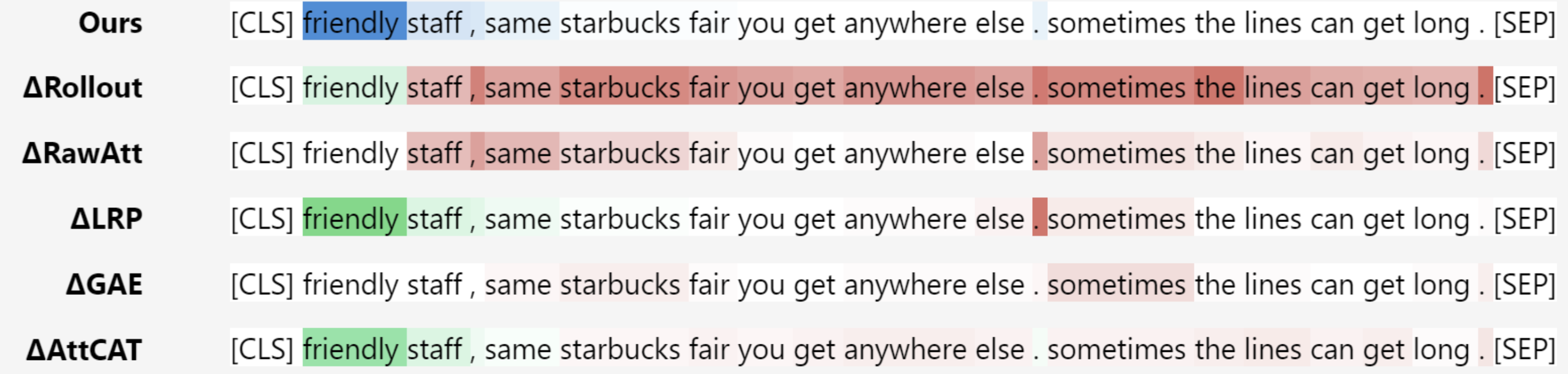}
    \caption{The comparison of attribution scores between our method (shown in the first line) and baselines on a \textbf{positive} classified comment in Yelp Polarity.}
    \label{fig:visu_yelp_1}
\end{figure*}
\section{Additional Visualization Results}
\label{apdx:extra_visul}
In this section, we provide visualization results of the attribution score difference in MNLI (Fig.~\ref{fig:visu_mnli_0}, \ref{fig:visu_mnli_1} and \ref{fig:visu_mnli_2}), IMDB (Fig.~\ref{fig:visu_imdb_0} and \ref{fig:visu_imdb_1}), and Yelp (Fig.~\ref{fig:visu_yelp_0} and \ref{fig:visu_yelp_1}), which include the task of classification of sentence pair and long text and each dataset, we randomly obtain a data from each class.
For all of the above figures, as we mentioned in Fig.~\ref{fig:visul}, tokens highlighted in green represent those receiving more attention from our method than the baseline, while those in red signify the opposite. Our method emphasizes more on both internal and interaction information.

\end{document}